\documentclass[10pt,twocolumn,letterpaper]{article}

\usepackage{wacv}              
\usepackage[dvipsnames]{xcolor}
\newcommand{\red}[1]{{\color{red}#1}}

\definecolor{cvprblue}{rgb}{0.21,0.49,0.74}
\usepackage{graphicx}
\usepackage{amsmath}
\usepackage{amssymb}
\usepackage{booktabs}
\usepackage{csquotes}
\usepackage{colortbl}
\usepackage{bm}
\usepackage{enumitem}
\usepackage[pagebackref,breaklinks,colorlinks,citecolor=cvprblue]{hyperref}
\usepackage{comment}
\usepackage[linesnumbered,ruled]{algorithm2e}

\usepackage[accsupp]{axessibility} 

\usepackage[capitalize]{cleveref}
\crefname{section}{Sec.}{Secs.}
\Crefname{section}{Section}{Sections}
\Crefname{table}{Table}{Tables}
\crefname{table}{Tab.}{Tabs.}

\def\wacvPaperID{*****} 
\def\confName{WACV}
\def\confYear{2025}

\title{Point-JEPA: A Joint Embedding Predictive Architecture for Self-Supervised Learning on Point Cloud}

\author{
    Ayumu Saito \quad
    Prachi Kudeshia \quad
    Jiju Poovvancheri \\
    Graphics and Spatial Computing Lab, Saint Mary's University, Halifax, Canada \\
    {\tt\small \{ayumu.saito, prachi.kudeshia, jiju.poovvancheri\}@smu.ca}
}

\begin{document}

\maketitle
\begin{abstract}
Recent advancements in self-supervised learning in the point cloud domain have demonstrated significant potential.
However, these methods often suffer from drawbacks such as lengthy pre-training time, the necessity of reconstruction in the input space, and the necessity of additional modalities.
In order to address these issues, we introduce Point-JEPA, a joint embedding predictive architecture designed specifically for point cloud data.
To this end, we introduce a sequencer that orders point cloud patch embeddings to efficiently compute and utilize their proximity based on their indices during target and context selection. 
The sequencer also allows shared computations of the patch embeddings' proximity between context and target selection, further improving the efficiency.
Experimentally, our method demonstrates state-of-the-art performance while avoiding the reconstruction in the input space or additional modality. In particular, Point-JEPA attains a classification accuracy of $\bm{93.7} \scriptstyle\pm 0.2$ \% for linear SVM on ModelNet40 
surpassing all other self-supervised models. Moreover, Point-JEPA also establishes new state-of-the-art performance levels across all four few-shot learning evaluation frameworks. The code is available at \url{https://github.com/Ayumu-J-S/Point-JEPA}
\end{abstract}     
\section{Introduction}
\label{sec:intro}
\begin{figure}
    \centering
    \includegraphics[width=8.5cm]{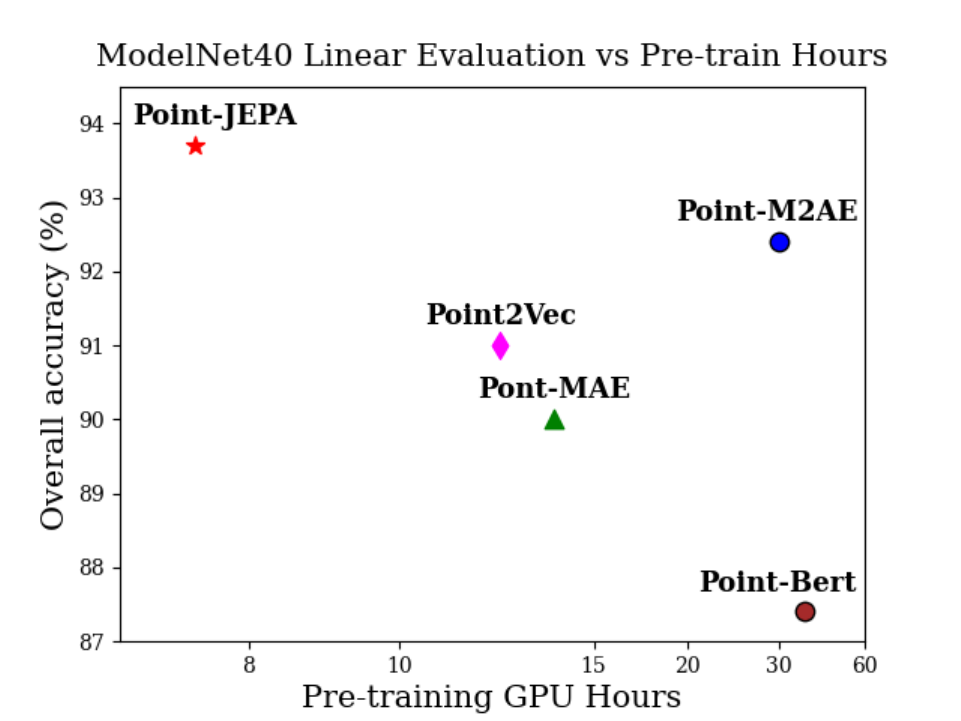}
    \caption{\textbf{ModelNet40 Linear Evaluation.}
    Pre-training time on NVIDIA RTX A5500 and overall accuracy with SVM linear classifier on ModelNet40 \cite{ModelNet}.
   We compare PointJEPA with previous methods utilizing standard Transformer architecture.}
    \label{fig:time}
\end{figure}
The growing accessibility of affordable consumer-grade 3D sensors has led to the widespread adoption of point clouds as a preferred data representation for capturing real-world environments. However, the existing point cloud understanding approaches \cite{guo2020deep} mostly rely on supervised training which requires time-consuming and labor-intensive manual annotations to semantically understand 3D environments. On the other hand, self-supervised learning (SSL) is an evolving paradigm that allows the model to learn a meaningful representation from unlabeled data. The success of self-supervised learning in advancing natural language processing and 2D computer vision has motivated its application in the point cloud domain for achieving state-of-the-art results on downstream tasks \cite{jaiswal2020survey}.
However, our initial investigation found that they require a significant amount of pre-training time as shown in \cref{fig:time}. The slow pre-training process can pose constraints in scaling to a larger dataset or complex and deeper models, hindering the key advantage of self-supervised learning; its capacity to learn a strong representation from a vast amount of data.\\
\indent The successful implementations of Joint-Embedding Predictive Architecture (JEPA) \cite{JEPA} for pre-training a model \cite{I-JEPA, bardes2023v} show JEPA's ability to learn strong semantic representations without the need for fine-tuning. The idea behind JEPA is to learn a representation by predicting the embedding of the input signal, called \textit{target}, from another compatible input signal, called \textit{context}, with the help of a predictor network. This allows learning in the representation space instead of the input space, leading to efficient learning. Inspired by I-JEPA \cite{I-JEPA}, we aim to apply Joint-Embedding Predictive Architecture in the point cloud domain, which introduces a promising direction for self-supervised learning in the point cloud understanding. However, unlike images, unordered point clouds pose a unique challenge to applying JEPA due to their inherently permutation-invariant nature.
The unordered nature of the point cloud data makes the context and target selection of the data difficult and inefficient, especially if we aim to select spatially contiguous patches similar to I-JEPA \cite{I-JEPA}. Therefore, we introduce Point-JEPA to overcome this challenge, while utilizing the full potential of Joint-Embedding Predictive Architecture for computational efficiency. Point-JEPA utilizes an efficient greedy sequencer to assist the model in selecting patch embeddings that are spatially adjacent. Our empirical studies indicate that Point-JEPA efficiently learns semantic representations from point cloud data with faster pre-training times compared to alternative state-of-the-art methods.
The specific contributions of this work are as follows. 
\begin{itemize}[noitemsep]
\item We present a Joint-Embedding Predictive Architecture, called Point-JEPA, for point cloud self-supervised learning. Point-JEPA efficiently learns a strong representation from point cloud
data without reconstruction in the input space or additional modality.
\item We propose a point cloud patch embedding ordering method for Joint-Embedding Predictive Architecture, utilizing a greedy algorithm based on spatial proximity. 
\end{itemize}

 \section{Related Work}
\label{sec:related}
Recent advancements in self-supervised learning in 2D computer vision \cite{I-JEPA, SimCLR, DINO, MAE, BYOL, jaiswal2020survey,wang2022self, mu2022slip} and natural language processing \cite{devlin2018bert, DBLP:journals/corr/abs-2005-14165, radford2019language, DBLP:journals/corr/abs-1910-10683} have inspired its application to point cloud processing. In this section, we review existing self-supervised learning methods in the point cloud domain and explore the concept of the Joint Embedding Predictive Architecture.
\subsection{Generative Learning}
Generative models learn representations by reconstructing the input signal within the same input space, capturing its underlying structure and features. 
For example, based on a popular NLP model Bert \cite{devlin2018bert}, Point-Bert \cite{Point-bert} introduces generative pretraining to the point cloud using a discrete variational autoencoder to transform the point cloud into discrete point tokens. However, this model heavily relies on data augmentation and suffers from the early leakage of location information, which makes pre-training steps relatively complicated and computationally expensive. 
To overcome this issue, Point-MAE \cite{Point-MAE} presents a lightweight, flexible, and computationally efficient solution by bypassing the tokenization and reconstructing the masked point cloud patches. 
On the other hand, PointGPT \cite{PointGPT} introduces an auto-regressive learning paradigm in the point cloud domain. 
Such generative pre-training in the point cloud domain learns a robust representation; however, it suffers from computational inefficiency due to the reconstruction of the data in the input space.
\subsection{Joint Embedding Architecture}
Joint Embedding Architectures map the input data into a shared latent space that contains similar embeddings for semantically similar instances. These networks utilize regularization strategies such as contrastive learning and self-distillation to learn meaningful representations. Contrastive learning generates embeddings that are close for positive pairs and distant for negative pairs. 
For example, Du \etal \cite{du2021self} introduces a contrastive learning approach that treats different parts of the same object as negative and positive examples.
\begin{figure*}[t]
    \centering
    \hspace*{0.7cm} 
    \includegraphics[width=0.9\linewidth]{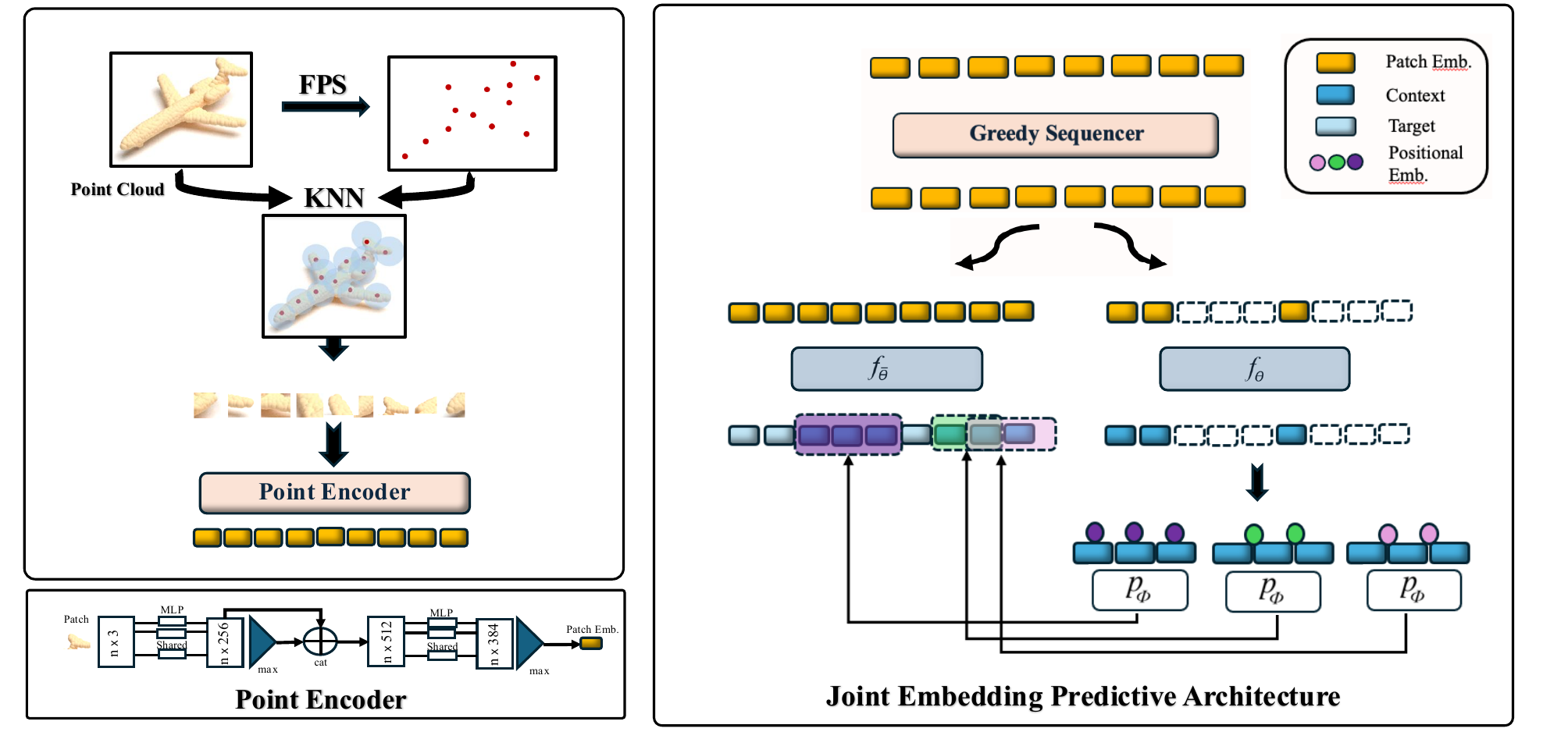}
    \caption{ 
    \textbf{Schematic renderings illustrating the process of creating embeddings. (Top left), point encoder (bottom left) and Point-JEPA (right).} Point cloud patches are generated using furthest point sampling (FPS)~\cite{fps} and $k$-nearest neighbor (KNN) methods, a mini PointNet (Point Encoder) is used to generate patch embeddings which are subsequently fed to the JEPA architecture. We use standard Transformer \cite{Transformer} architecture for context ($f_{\theta}$) and target ($f_{\overline{\theta}}$) encoders as well as predictor ($p_{\phi}$). }
    \label{fig:point-jepa}
\end{figure*}
Unlike contrastive learning, a self-distillation network employs two identical networks with distinct parameters, commonly known as the teacher and student, where the teacher guides the student by providing its predictions as targets. For example, 
in Point2Vec \cite{Point2Vec}, the teacher receives the patches of point clouds while the student receives a subset of these patches. Further, a shallow Transformer 
learns meaningful and robust representation from the masked positional information and the contextualized embedding from the partial-view input. In self-distillation networks, no reconstruction in the input space results in faster training than in generative models. However, as shown in \cref{fig:time}, it requires longer training to learn a meaningful representation. On the other hand, contrastive learning excels in performance; however, its effectiveness highly depends on the careful selection of positive and negative samples as well as the data augmentation techniques to ensure transferable representations for downstream tasks \cite{jaiswal2020survey}.
\subsection{Joint Embedding Predictive Architecture (JEPA)}
A self-supervised learning architecture JEPA \cite{oord2018representation} learns representation using a predictor network that predicts one set of encoded signal $y$ based on another set of encoded signal $x$, along with a conditional variable $z$ that controls the prediction. In the predictor network, encoders initially process both the target and the context signals to represent them in embedding space. Conceptually JEPA has a large similarity to generative models which are designed to reconstruct masked part of the input. However, instead of directly operating on the input space, JEPA makes predictions in the embedding space. This allows the elimination of unnecessary input details to focus on learning meaningful representations. As a result, the model can abstract and represent the data more efficiently. Closely related to our work, the specific application of the architecture in the image domain can be seen in I-JEPA \cite{I-JEPA}. In this work, the context signal is created by selecting a block of patches while the target signals are created by sampling the rest of unselected patches. Experiments show faster convergence of I-JEPA to learn highly semantic representation. Therefore, to ensure faster pretraining in self-supervised learning for point cloud understanding, we aim to apply JEPA on point cloud data.

\section{Point-JEPA Architecture}
In this section, we describe our JEPA architecture for pretraining in the point cloud domain. Our goal is to adapt JEPA~\cite{JEPA} for use with point cloud data while evaluating its performance and implementation efficiency. The overall framework, as shown in \cref{fig:point-jepa}, first converts the point cloud to a set of patch embeddings, then a greedy sequencer arranges them in sequence based on their spatial proximity to each other, and Joint-Embedding Predictive Architecture is applied to the ordered patch embeddings. We utilize a mini PointNet \cite{PointNet} architecture for encoding the grouped points and standard Transformer \cite{Transformer} architecture for the context and target encoder as well as the predictor. It is important to note that our JEPA architecture operates on embeddings instead of patches in order to share the point encoder network between context and target encoder for efficiency similar to Point2Vec \cite{Point2Vec}.  
\subsection{Point Cloud Patch Embedding}
Building on previous studies that utilize the standard Transformer architecture for point cloud objects \cite{Point-MAE, PointGPT, Point2Vec}, we adopt a process that embeds groups of points into patch embeddings.
Given a point cloud object, $P\subset \mathbb{R}^3$ consisting of $n$ points, $c$ center points are first sampled using the farthest point sampling \cite{fps}. 
Then we employ the $k$-nearest neighbors algorithm to identify and select the $k$ closest points surrounding each of the $c$ designated center points. 
These point patches are then normalized by subtracting the center point coordinates from the coordinates of the points in the patches.
This allows the separation between local structural information and the positional information of the patches. 
In order to embed the local point patches, we utilize a mini PointNet \cite{PointNet} architecture.
This ensures that the patch embedding remains invariant to any permutations of data feeding order of points within the patch.
Specifically, this PointNet contains two sets of a shared multi-layer perceptron (MLP)  and a max-pooling layer as shown in Fig. \ref{fig:point-jepa}. 
First, a shared MLP  maps each point into a feature vector. Then, we apply max-pooling to these vectors and concatenate the result back to the original feature vector.
Subsequently, a shared MLP processes these concatenated vectors, followed by a max-pooling operation to generate a set of patch embeddings $T$ of $P$.
\begin{algorithm}
    \SetKwInOut{Input}{Input}
    \SetKwInOut{Output}{Output}

    \Input{Set of patch emb., $T = \{t_1, t_2,..,t_r\}$}
    \Output{Set of spatially contiguous patch emb., $T'$}
    Find the initial patch emb. $t = minCoordSum(T)$\;
    Set $T' =\{t\}$\;
    $T = T\setminus\{t\}$\;
    Initialize $prev\_t = t$\;
    \While{$T \neq \emptyset$}{
    Set $closest = \infty$ \;    
    \For {$t_i \in T$}
    { 
    $dis = \parallel center(prev\_t)-center(t_i)\parallel $\;
    \If{$dis \leq closest$}
      {        
        Set $closest = dis $\;
        Set $index = i$\;
      }      
      }
      $T' = T'\cup \{t\_index\}$\;
      $T = T\setminus\{t\_index\}$\;
      $prev\_t = t\_index$\;
      }    
    \caption{Greedy sequencer strategy}
    \label{alg:gts}
\end{algorithm}

\subsection{Greedy Sequencer}
Due to the previously observed benefits of having targets and context clustered together in close spatial proximity, a configuration known as a block in I-JEPA \cite{I-JEPA}, we aim to sample patch embeddings that are spatially close to each other.
As previously mentioned, point cloud data is permutation invariant to data feeding order, which implies that even if the indices of patch embeddings are sequential, they might not be spatially adjacent.
Furthermore, our approach involves the selection of $M$ spatially contiguous blocks of encoded embeddings as the target while ensuring that the context does not include the patch embeddings corresponding to these embedding vectors (details in the next paragraph).
To address these challenges, we apply a greedy sequencer that is applied after producing patch embeddings similar to z-ordering in PointGPT \cite{PointGPT}. This sequencer orders patch embeddings based on their associated center points ( \cref{alg:gts}). 
The process is initiated by selecting the center point with the lowest sum of coordinates ($minCoorSum(T)$) as the starting point, along with its associated patch embedding.  In each subsequent step, the center point closest to the one previously chosen and its associated patch embedding are selected.
This is iterated until the sequencer visits all of the center points. The resulting arrangement of patch embeddings ($T'=\{t_1^{'}, t_2^{'},...,t_r^{'}\}$) is in a sequence where contiguous elements are also spatially contiguous in most cases.
This allows the shared computation of spatial proximity between context and target selection. At the same time, this also allows simpler implementation for context and target selection.
It is worth noting, however, that selecting two adjacent patch embedding indices in this setting does not always guarantee spatial proximity; there might be a gap between them. While this is true, the experiment results show that this iterative ordering is effective enough in our JEPA architecture. Additionally, this rather simple approach is parallelized across batches, making it more efficient for large datasets or point clouds. 
Not only can we compute pairwise distances for all points within a batch in a single forward pass but also run the iterative process of simultaneously selecting the next closest point across the batch.
This enables faster computation on modern GPUs, ensuring that the nearest points are selected efficiently while keeping the algorithm feasible even for large batch sizes.

\subsection{JEPA Components}
\paragraph{Context and Target}
Targets in Point-JEPA can be considered patch-level representations of the point cloud object, which the predictor aims to predict.
As illustrated in \cref{fig:point-jepa}, the target encoder initially encodes the patch embedding conventionally, and we randomly select $M$ possibly overlapping target blocks, which are sets of adjacent encoded embeddings.
Specifically, we define  
   $y(i) = {\{y_j\}}_{j \in B_i}$
as the $i^{\text{th}}$ target representation block, where $B_i$ denotes the set of mask indices for the $i^{\text{th}}$ block. Here, we denote the encoded embeddings as $y = \{y_1, y_2, \ldots y_{n} \}$, where $y_k = f_{\overline{\theta}}(t_k^{'})$ is the representation associated with the $k^{th}$ centre point. 
It is important to note that masking for the target is applied to the embedding vectors derived from the patch embeddings that have passed through the transformer encoder $f_{\overline{\theta}}$. This ensures a
high semantic level for the target representations. 
\begin{figure}[h]
    \centering
    \includegraphics[width=0.9\linewidth]{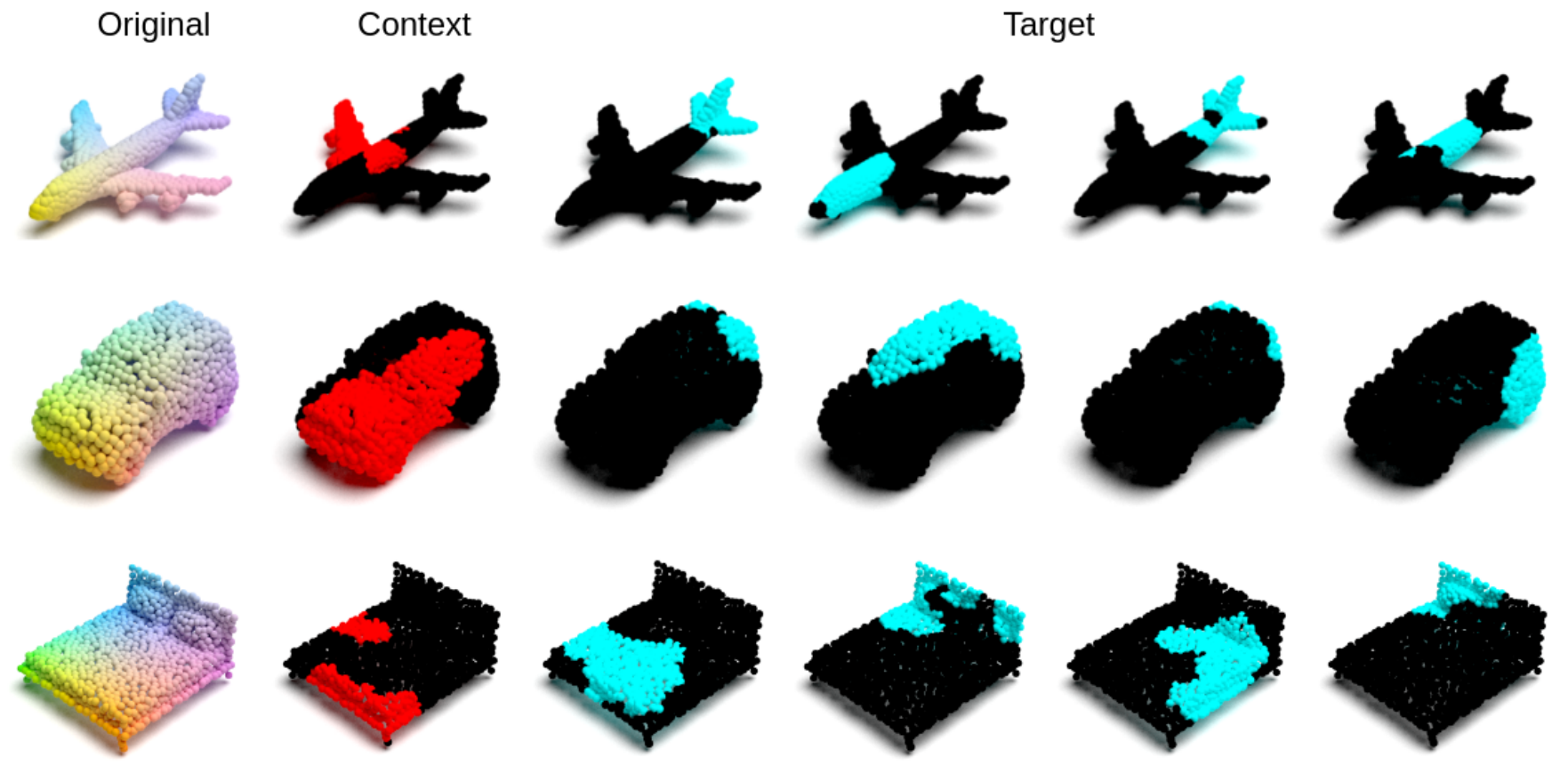}
    \caption{\textbf{Context and Targets.}
    We visualize the corresponding grouped points of context and target blocks. 
    Here, we use (0.15, 0.2) for the target selection ratio and (0.4, 0.75) for the context selection ratio. }
    \label{fig:enter-label}
\end{figure}

Context, on the other hand, is the representation of the point cloud object which is passed to the predictor to facilitate the reconstruction of target blocks. Unlike target blocks, masking is applied to the patch embeddings during the creation of context blocks. 
This allows the context-encoder $f_{\theta}$ to represent the uncertainties in the object's representations when certain parts are masked.
Specifically, we first select a subset of patch embeddings $\hat{T} \subseteq T'$ that are spatially contiguous.
These selected embeddings are then fed to the context-encoder $f_{\theta}$ to generate a context block $x = \{x_j\}_{j \in B_x}$.
To prevent trivial learning, we also ensure that the indices of patch embeddings chosen for the context differ from those for the targets. Furthermore, the patch embeddings $T'$  are sorted such that embeddings that are adjacent in the data feeding order are also spatially close. This organization simplifies the selection of contiguous target and context blocks, despite the aforementioned complexities of point cloud data representation.

\paragraph{Predictor}
The task of predictor $p_{\phi}$ given targets $y$ and context $x$ is analogous to the task of supervised prediction.
Given a context as input $x$ along with a certain condition, it aims to predict the target representations $y$.
Here, the condition involves the mask tokens, which are created from shared learned parameters, as well as positional encoding, created from centre points associated with the targets.
That is
\begin{equation}
    \hat{y}(i) = \{\hat{y}_j\}_{j \in B_i} = p_{\phi}\left( x, \{m_j\}_{j \in B_i} \right)
\end{equation}
where $p_{\phi}\left( \cdot,\cdot \right)$ denotes the predictor and $\{m_j\}_{j \in B_i}$ denotes the mask token created from shared learnable parameter and positional encoding created from centre points.

\paragraph{Loss}
Because the predictor's task is to predict the representation produced by the target encoder, the loss can be defined to minimize the disagreement between the predictions and targets as follows.
\begin{equation}
    \frac{1}{M} \sum_{i=1}^{M} D(\hat{y}(i), y(i))
     = \frac{1}{M} \sum_{i=1}^{M} \sum_{j \in B_i} 
     \mathcal{L}(\hat{y}_j, y_j)
\end{equation}
Similar to Point2Vec \cite{Point2Vec}, we utilize Smooth L1 loss to measure the dissimilarity between each corresponding element of the target and predicted embedding because of its ability to be less sensitive to the outliers. 

\paragraph{Parameter Update}
We utilize AdamW \cite{AdamW} optimizer with cosine learning decay \cite{SGDR}. The target encoder and context encoder initially have identical parameters. The context encoder's parameters are updated via backpropagation, while the target encoders' parameters are updated using the exponential moving average of the context encoder parameters, that is $\overline{\theta} \xleftarrow{} \tau \overline{\theta} + (1 - \tau)\theta$
where $\tau \in [0,1]$ denotes the decay rate. 



\section{Experiments}

\begin{table}[ht]
  \centering
   \caption[Linear Evaluation on ModelNet40]{
        \textbf{Linear Evaluation on ModelNet40 \cite{ModelNet}.} 
        We compare Point-JEPA to self-supervised learning methods pre-trained on ShapeNet \cite{Shapenet}. * signifies the linear evaluation results as indicated in \cite{Point-M2AE, I2P-MAE}. ** signifies results with Transformer backbone.
  }
  \label{tab:linear_eval}
  \begin{tabular}{lc}
    \toprule
    methods & \multicolumn{1}{c}{Overall Accuracy} \\
    \midrule
    Latent-GAN  \cite{Latent-GAN}      & 85.7 \\
    3D-PointCapsNet \cite{3DCapsNet}   & 88.9\\
    STRL \cite{STRL}                   & 90.3\\
    Sauder \etal  \cite{Saunder2021}   &  90.6 \\
    Fu \etal \cite{fu2022distillation}
                                       & 91.4\\
    Transformer-OcCo* \cite{Point-bert}  &  89.6 \\
    Point-BERT* \cite{Point-bert}        &  87.4 \\
    Point-MAE* \cite{Point-MAE} & 90.0 \\
    Point-M2AE \cite{Point-M2AE}        & 92.9  \\
    CluRender** \cite{mei2024unsupervised} & 93.2  \\
\arrayrulecolor{black!10}\midrule\arrayrulecolor{black}
    Point-JEPA (Ours)         & $\bm{93.7} \scriptstyle\pm 0.2$  \\
    \bottomrule
  \end{tabular}
\end{table}

In this section, we first describe the details of self-supervised pre-training. Further, we compare the performance of the learned representation to the state-of-the-art self-supervised learning methods in the point cloud domain that utilizes the ShapeNet \cite{Shapenet} dataset in pre-training. We specifically evaluate the learned representation using linear probing, end-to-end fine-tuning, and a few-shot learning setting. Finally, ablation experiments are conducted to gain insights into the principal characteristics of Point-JEPA.
\subsection{Self-Supervised Pre-training}
We pre-train our model on training set of ShapeNet \cite{Shapenet} following the previous studies utilizing the standard Transformer \cite{Transformer} architecture such as Point-MAE \cite{Point-MAE}, Point-M2AE \cite{Point-M2AE}, PointGPT \cite{PointGPT}, and Point2Vec \cite{Point2Vec} for the fair comparison.
The dataset consists of 41952 3D point cloud instances created from synthetic 3D meshes from 55 categories. The standard Transformer \cite{Transformer} architecture is used for the context and target encoder as well as the predictor. During pre-training, we set the number of center points to 64 and the group size to 32. The point tokenization is applied to the input point cloud containing 1024 points per object. We set the depth of the Transformer in the context and target encoder to 12 with the embedding width of 384 and 6 heads. For the predictor, we use the narrower dimension of 192 following I-JEPA \cite{I-JEPA}. The depth of the predictor is set to 6, and the number of heads is set to 6. Our experiments are conducted on NVIDIA RTX A5500 and NVIDIA A100 SXM4.
We note that our method only takes 7.5 hours on RTX A5500 for pretraining (see Fig. \ref{fig:time}) which is less than half of that of PointM2AE \cite{Point-M2AE} and about 60\% of Point2Vec \cite{Point2Vec} time requirement for pretraining. Adhering to the standard convention, we use overall accuracy for classification tasks and mean IoU for part segmentation tasks.

\begin{table*}[ht]
  \centering
  \caption[Result of Fine-Tuning]{\textbf{End-to-End Classification.}
    Overall accuracy on ModelNet40 \cite{ModelNet} and ScanObjNN \cite{ScanObj} with end-to-end fine-tuning. We specifically compare our methods to the method utilizing standard
    Transformer architecture pre-trained on ShapeNet \cite{Shapenet} with only point cloud (no additional modality).
  }
  \label{tab:e2e-class}
    \setlength\tabcolsep{4pt} 
  \begin{tabular}{ccccccccc}
    \toprule
     & &\multicolumn{7}{c}{Overall Accuracy} \\
    \cmidrule(lr){3-9} 
    Method & Reference & \multicolumn{3}{c}{\textbf{ModelNet40}} &  \multicolumn{4}{c}{\textbf{ScanObjNN}} \\ 
    \cmidrule(lr){3-5}\cmidrule(lr){6-9}
     & & \#Points & +Voting & -Voting & \#Points&\texttt{OBJ-BG} & \texttt{OBJ-ONLY} & \texttt{OBJ-T50-RS}\\
    \midrule
    Point-BERT \cite{Point-bert} &CVPR2022 & 1k  & 93.2 & 92.7 & 1k &87.4 & 88.1 & 83.1 \\
    Point-MAE \cite{Point-MAE}   &ECCV2022 & 1k& 93.8  & 93.2 &2k &90.0 & 88.3 & 85.2 \\
    Point-M2AE \cite{Point-M2AE}& NeurIPS2022& 1k & 94.0  & 93.4 & 2k &91.2 & 88.8 & 86.4 \\
    Point2Vec \cite{Point2Vec}& GCPR2023& 1k & \textbf{94.8} & \textbf{94.7} & 2k &91.2 & 90.4 & 87.5 \\
    PointGPT-S \cite{PointGPT}& NeurIPS2023& 1k  & 94.0 & -- & 2k &91.6 & 90.0 & 86.9 \\
    PointDiff \cite{zheng2024point}& CVPR2024 & --  & -- & -- & 1k &\textbf{93.2} & \textbf{91.9} & \textbf{87.6} \\
    \arrayrulecolor{black!10}\midrule\arrayrulecolor{black}
    Point-JEPA (Ours)& -     
        & 1k&$94.1 \scriptstyle\pm 0.1$  & 
        $ 93.8 \scriptstyle\pm 0.2 $ & 2k &92.9 $\scriptstyle\pm 0.4$ & $90.1 \pm 0.2$ & $	86.6 \pm 0.3$ \\
    \bottomrule
  \end{tabular}
\end{table*}

\subsection{Downstream Tasks}
In this section, we report the performance of the learned representation on several downstream tasks.
Following the previous studies \cite{Point-bert, Point-M2AE, PointGPT, Point2Vec}, we report the overall accuracy as a percentage. To account for variability across independent runs, we report the mean accuracy and standard deviation from 10 independent runs with different seeds, unless specified otherwise.  

\paragraph{Linear Probing.}
After pre-training on ShapeNet \cite{Shapenet}, we evaluate the learned representation via linear probing on ModelNet40 \cite{ModelNet}. 
Specifically, we freeze the learned context encoder and place the SVM classifier on top. To enforce invariance to geometric transformation, we utilize max and mean pooling on the output of the Transformer encoder \cite{Point-MAE, Point2Vec}. We utilize 1024 points for both training and test sets. As shown in \cref{tab:linear_eval}, our method achieves state-of-the-art accuracy, providing $+0.8\%$ performance gain, showing the robustness of the learned representation. 

\begin{table}[b]
  \caption[Few-Shot classification on ModelNet40]{
  \textbf{Result of Few-Shot classification on ModelNet40 \cite{ModelNet}.}
  10 independent trials are completed under one setting.
  We report mean and standard deviation over $10$ trials. ** signifies results with Transformer backbone.}
  \label{tab:modelnet_fewshot_results}
  \setlength\tabcolsep{2pt}
  \begin{tabular}{lcccc}
    \toprule
     & \multicolumn{4}{c}{Overall Accuracy} \\
    \cmidrule(lr){2-5} 
     Method & \multicolumn{2}{c}{5-way} &  \multicolumn{2}{c}{10-way} \\
    \cmidrule(lr){2-3}\cmidrule(lr){4-5}
     & 10-shot & 20-shot & 10-shot & 20-shot\\
    \midrule
    Point-BERT \cite{Point-bert}  
        & $94.6 \scriptstyle\scriptstyle\pm 3.1$ & $96.3 \scriptstyle\pm 2.7$ & $91.0 \scriptstyle\pm 5.4$ & $92.7 \scriptstyle\pm 5.1$ \\
    Point-MAE \cite{Point-MAE} 
        & $96.3 \scriptstyle\pm 2.5$ & $97.8 \scriptstyle\pm 1.8$ & $92.6 \scriptstyle\pm 4.1$ & $95.0 \scriptstyle\pm 3.0$ \\
    Point-M2AE \cite{Point-M2AE} 
        & $96.8 \scriptstyle\pm 1.8$ & $98.3 \scriptstyle\pm 1.4$ & $92.3 \scriptstyle\pm 4.5$ & $95.0 \scriptstyle\pm 3.0$ \\
    Point2Vec \cite{Point2Vec}
       & $97.0 \scriptstyle\pm 2.8 $ & $98.7 \scriptstyle\pm 1.2$ & $93.9 \scriptstyle\pm 4.1 $ & $95.8 \scriptstyle\pm 3.1$ \\
    PointGPT-S \cite{PointGPT}
       & $96.8 \scriptstyle\pm 2.0 $ & $98.6 \scriptstyle\pm 1.1$ & $92.6 \scriptstyle\pm 4.6$ & $95.2 \scriptstyle\pm 3.4$ \\
    CluRender** \cite{mei2024unsupervised}
        & $97.2 \scriptstyle\pm 2.3 $ & $98.4 \scriptstyle\pm 1.3$ & $93.7 \scriptstyle\pm 4.0$ & $96.0 \scriptstyle\pm 2.9$ \\
\arrayrulecolor{black!10}\midrule\arrayrulecolor{black}
    Point-JEPA (Ours) 
        & $\bm{97.4} \scriptstyle\pm 2.2$   & $\bm{99.2} \scriptstyle\pm  0.8 $ & $\bm{95.0} \scriptstyle\pm 3.6 $  & $\bm{96.4} \scriptstyle\pm 2.7 $ \\ 
    \bottomrule
  \end{tabular}
\end{table}

\paragraph{Few-Shot Learning}
We conduct few-shot learning experiments on Modelnet40 \cite{ModelNet} in $m$-way, $n$-shot setting as shown in  \cref{tab:modelnet_fewshot_results}.
Specifically, we randomly sample $n$ instances of $m$ classes for training and select 20 instances of $m$ support classes for evaluation. For one setting, we run 10 independent runs under a fixed random seed on 10 different folds of dataset and report mean and standard deviation of overall accuracy.
As shown in  \cref{tab:modelnet_fewshot_results}, our method exceeds the performance of current state-of-the-art in all settings and yields a $+1.1\%$ improvement in the most difficult 10-way 10-shot setting, showing the robustness of the learned representation of Point-JEPA, especially in the low-data regime.

\paragraph{End-to-end Fine-Tuning}
We also investigate the performance of the learned representation via end-to-end fine-tuning.
After pre-training, we utilize the context encoder to extract the max and average pooled outputs.
These outputs are then processed by a three-layer MLP for classification tasks.
This class-specific head as well as the context encoder is fine-tuned end-to-end on ModelNet40 \cite{ModelNet} and ScanObjectNN \cite{ScanObj}.
ModelNet40 consists of 12311 synthetic 3D objects from 40 distinct categories, while ScanObjectNN contains objects from 15 classes, each containing 2902 unique instances collected by scanning real-world objects.
For ModelNet40, we sub-sample 1024 points per object and sample 64 center points with 32 points in each point patch.
On the other hand, we utilize all 2048 points for the ScanObjNN dataset and sample 128 center points with 32 nearest neighbors for the grouped points.
\begin{table*}[ht]
        \caption[Sampling Methods]{
        \textbf{Masking Strategies.}
            Multi-block and single-block masking strategies and their effect on the learned representation.
        }
        \centering
        \normalsize{
            \begin{tabular}{l l l l l c}
            \toprule
            \multicolumn{3}{l}{\bf Targets} & \multicolumn{2}{l}{\bf Context} & \bf OA\\
            \cmidrule(lr){1-3}\cmidrule(lr){4-5}\cmidrule(lr){6-6}
            Strategy & Ratio & Freq. & Strategy & Ratio & Modelnet40 Linear \\
            \toprule
            random      & (0.6, 0.6)  &  1 & rest & -- & 92.5 \\
            contiguous  & (0.6, 0.6)  &  1 & rest & -- &  92.3 \\
            contiguous  & (0.15, 0.2) &  4 & contiguous & (0.4, 0.75) & \textbf{93.7} \\
            \bottomrule
            \end{tabular}
        }
        \label{tab:maskig_strategy}
\end{table*}
As shown in \cref{tab:e2e-class}, our method achieves competitive results when compared to other state-of-the-art methods.
Especially, in the \texttt{OBJ-BG} variant of the ScanObjNN \cite{ScanObj} dataset, which presents a realistic representation of a point cloud that includes both the object and its background, our method achieves the overall accuracy marginally lower than PointDiff \cite{zheng2024point} while an improvement of $+1\%$ over other SSL methods.
This shows the learned representation obtained from pre-training with Point-JEPA can easily be transferred to a classification task.
\begin{table}[t]
  \centering
  \caption[Result of Part Segmentation on ShapeNetPart]{
  \textbf{Part Segmentation on ShapeNetPart \cite{Shapenet}}. mIoU$_C$ is the mean IoU for all part categories, and mIoU$_I$ is the mean IoU for all instances.  
  }
  \label{tab:segmentation}
  \begin{tabular}{lcc}
    \toprule
    Method & mIoU$_C$ & mIoU$_I$ \\
    \midrule
    Transformer-OcCo \cite{Point-bert} & 83.4 & 85.1 \\
    Point-BERT \cite{Point-bert} & 84.1 & 85.6 \\
    Point-MAE \cite{Point-MAE} & 84.1 & 86.1 \\
    Point-M2AE \cite{Point-M2AE} & \textbf{84.9} & \textbf{86.5} \\
    Point2Vec \cite{Point2Vec} & 84.6 & 86.3 \\
    PointGPT-S \cite{PointGPT} & 84.1 & 86.2 \\
    \midrule
    Point-JEPA (Ours) & $83.9\pm0.1$ & $85.8\pm0.1$ \\
    \bottomrule
  \end{tabular}
\end{table}
\paragraph{Part Segmentation}
Following previous studies \cite{Point-bert, Point-M2AE, Point2Vec, Point-MAE, PointGPT}, we report the performance of Point-JEPA in part segmentation task.
Here, we utilize the ShapeNetPart \cite{Shapenet} dataset, consisting of 16881 objects from 16 categories.
We utilize the identical architecture employed in Point2Vec \cite{Point2Vec} for this task.
Specifically, we take the embeddings from $4^{\text{th}}$, $8^{\text{th}}$, and $12^{\text{th}}$ Transformer block and take the average of them.
Then, we apply mean and average pooling to this averaged output. 
The max and mean pooled embedding along with a one-hot encoded class label of an object is used as a global feature vector for the object.
The original averaged output is also up-sampled using the PointNet++ \cite{PointNet++} feature propagation layer to create a feature vector for each point.
Then, each feature vector is concatenated with the global feature vector. A shared MLP is utilized on this concatenated vector to predict the segmentation label for the given point.
Although the Point-JEPA shows competitive results, as shown in \cref{tab:segmentation}, its performance is slightly worse than the state-of-the-art methods.
\paragraph{Limitations}
Point-JEPA's comparatively weaker performance in segmentation and its superior learned representation in classification indicate that the proposed approach emphasizes global features over local features. Additionally, the effectiveness of Point-JEPA for processing larger point clouds is uncertain due to redundancy in many areas in data and requires further study.

\subsection{Ablation Study}
We conducted thorough ablation studies to understand the effect of moving parts of Point-JEPA. We pre-train Point-JEPA on the ShapeNet \cite{Shapenet} dataset under various settings and evaluate the learned representation with linear probing on the ModelNet40 \cite{ModelNet} dataset. 
\paragraph{Masking Strategy.}
We investigate the impact of the masking type on the performance. We consider single-block masking and multi-block masking. For single-block masking strategies, we consider random masking and contiguous masking.  For random masking, we randomly select the 60\% of indices out of all encoded embedding vectors. Similarly, for contiguous masking, embedding vectors that are spatially contiguous are selected. In this setting, all patch embeddings not corresponding to the selected target blocks are used as context (denoted as rest). On the other hand, in the multi-block masking setting, we sample multiple possibly overlapping spatially contiguous embedding vectors as targets, and we remove the corresponding patch embeddings already selected for targets during context selection. In this setting, we set the ratio range of 0.15 to 0.2 for targets, while we set the ratio range of 0.4 to 0.75 for context.  As shown in \cref{tab:maskig_strategy}, the single-block masking achieves sub-optimal performance regardless of the spatial contiguity of the target embedding. It shows that our method learns stronger representation by utilizing a smaller amount of targets with a larger frequency.

\begin{table}[htb]
        \caption[Initial Point]{
         Experiment with different sequencers using the ModelNet40 linear evaluation and their corresponding training times on \emph{RTX 5500 GPU} for 500 epochs. 
        }
        \centering
        \normalsize{
             \setlength{\tabcolsep}{3pt}
            \begin{tabular}{l c c}
            \toprule
            Sequencer &  Modelnet40 Linear & Train Time \\
            \midrule  
            z-ordering~\cite{morton}              & 93.4 & 8.30h\\
            Hilbert ordering \cite{hilbert-program}  & 91.8 & 10.78h\\
            Greedy (min index)            & 92.7  & 7.47h \\
            Greedy (min coordinate)   & \textbf{93.7 } & 7.47h \\
            \bottomrule
            \end{tabular}
        }
        \label{tab:sequencer_inital_point}
\end{table}

\paragraph{Sequencer}
\label{parag:sequencer_ab}
At the heart of employing Point-JEPA for point cloud processing is the ability to translate point cloud data into a sequence of spatially contiguous patch embeddings. As shown in \cref{tab:sequencer_inital_point}, we conduct the ablation experiment with different sequencer strategies including both the \textit{$z$-order}~\cite{morton} and \textit{Hilbert-order}~\cite{hilbert-program} space-filling curves, as well as the proposed \textit{greedy sequencer}. The \textit{greedy sequencer} is evaluated in two versions: one starting with the point that has the minimum index in data feeding order, and the other starting with the point that has the minimum coordinate sum.
The \emph{greedy sequencer} with the minimum coordinate approach exhibits an overall accuracy improvement compared to all the other algorithms.
Additionally, the \emph{greedy sequencer} offers computational efficiency over \emph{z-order} and \emph{Hilbert-order} due to its ability to parallelize computations, as mentioned previously. 
Notably, the greedy sequencer with minimum coordinate sum initial point guarantees that the starting center point in the sequencer lies near the edge of the object. As shown in \cref{tab:sequencer_inital_point}, selecting the point near the edge of the object as the starting point helps the model learn a stronger representation.

\begin{figure}[htbp]
    \centering
    \includegraphics[width=8.5cm]{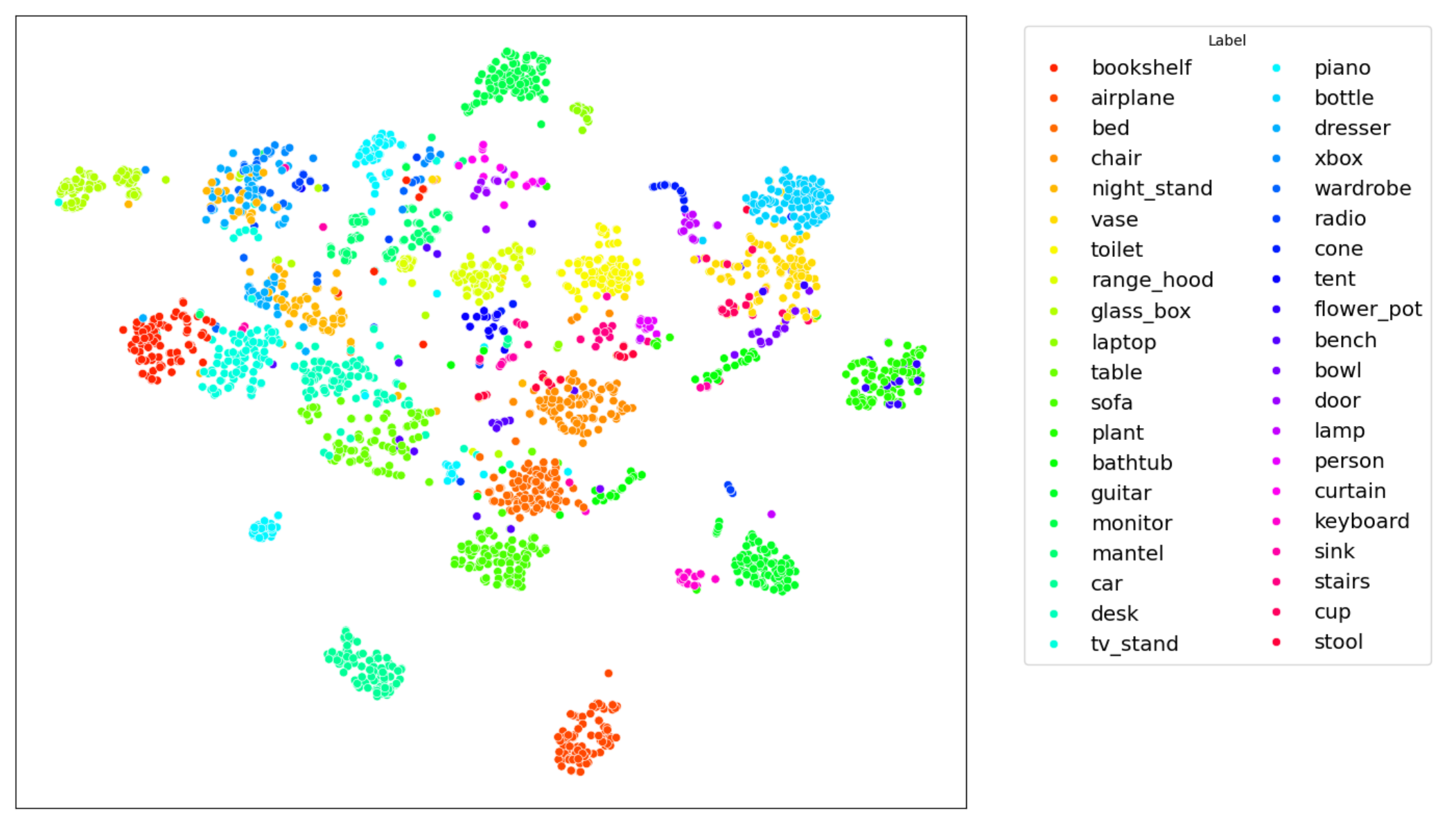}
    \caption{
    \textbf{Embedding Visualization on ModelNet40\cite{ModelNet}.}
    We visualize the context encoder's learned representation with t-SNE \cite{tsne}.}
    \label{fig:embed_vis}
\end{figure}

\paragraph{Number of Target Blocks.}
We also consider the effect of the number of blocks chosen for targets on the performance of the learned representation while we keep the ratio for targets and context fixed.
As shown in \cref{tab:num_targets}, the performance increases as we increase the number of targets.
However, the performance decreases as you increase the number of target blocks after a specific frequency. 
We observe that our method benefits from having a sufficient amount of patch embeddings available for context encoding.
\begin{table}[!ht]
        \caption[Number of Target Blocks]{
        \textbf{Number of Target blocks.}
        We change the number (frequency) of target blocks while keeping the other components fixed.
        }
        \centering
        \normalsize{
            \begin{tabular}{l l l c}
            \toprule
            \multicolumn{2}{l}{\bf Targets} & \bf Context & \bf OA \\
            \cmidrule(lr){1-2}\cmidrule(lr){3-3}\cmidrule(lr){4-4}
            Ratio & Freq. & Ratio &  Modelnet40 Linear \\
            \toprule
            (0.15, 0.2) &  1 & (0.85, 1.0) &  93.0 \\
            (0.15, 0.2) &  2 & (0.85, 1.0) &  93.5 \\
            (0.15, 0.2) &  3 & (0.4, 0.75) &  93.4 \\
            (0.15, 0.2) &  4 & (0.4, 0.75) & \textbf{93.7} \\
            (0.15, 0.2) &  5 & (0.4, 0.75) & 93.4 \\
            (0.15, 0.2) &  6 & (0.4, 0.75) & 93.2  \\
                        \bottomrule

            \end{tabular}
        }
        \label{tab:num_targets}
\end{table}

\subsection{Visualization}
To qualitatively analyze the learned representation, we reduce the dimension of the learned representation by utilizing t-SNE \cite{tsne}. We introduce max and mean pooling on the output of the context encoder, similar to the classification setup, and apply t-SNE on the pooled embedding. We visualize the learned representation on ModelNet40 \cite{ModelNet} with no fine-tuning on the dataset. 
Despite being trained on the dataset, our context encoder produces discriminative features as shown in \cref{fig:embed_vis}, showing the robustness of the learned representation.
\section{Conclusion}
This work introduced Point-JEPA, a joint embedding predictive architecture applied to point cloud objects.
In order to efficiently select targets and context blocks even under the invariance property of point cloud data, we introduced a sequencer, which orders the center points and their corresponding patch embeddings by iteratively selecting the next closest center point.
This eliminates the necessity of computing spatial proximity between every pair of patch embeddings or encoded embeddings when sampling the targets and context.
Point-JEPA achieves state-of-the-art performance in downstream tasks, excelling in few-shot learning and linear evaluation.
This makes Point-JEPA highly useful when there is a large amount of unlabeled data and a limited amount of labeled data.
It is also worth noting that Point-JEPA converges much faster during pre-training, offering a more efficient pre-training alternative in the point cloud domain. Future work includes extending Point-JEPA for other downstream tasks such as object detection and scene level segmentation, and pre-training on large unlabeled hybrid datasets~\cite{PointGPT}. Additionally, the ability of JEPA to reduce dependency on large labeled datasets presents potential avenues for generating temporal predictive embeddings that anticipate point cloud evolution, thereby enhancing tasks like motion prediction, dynamic scene understanding, and anomaly detection.

{\small
\bibliographystyle{ieee_fullname}
\bibliography{main}

\begin{thebibliography}{10}\itemsep=-1pt

\bibitem{Latent-GAN}
Panos Achlioptas, Olga Diamanti, Ioannis Mitliagkas, and Leonidas~J. Guibas.
\newblock Representation learning and adversarial generation of 3d point clouds.
\newblock {\em CoRR}, abs/1707.02392, 2017.

\bibitem{I-JEPA}
Mahmoud Assran, Quentin Duval, Ishan Misra, Piotr Bojanowski, Pascal Vincent, Michael Rabbat, Yann LeCun, and Nicolas Ballas.
\newblock Self-supervised learning from images with a joint-embedding predictive architecture.
\newblock In {\em Proceedings of the IEEE/CVF Conference on Computer Vision and Pattern Recognition}, pages 15619--15629, 2023.

\bibitem{bardes2023v}
Adrien Bardes, Quentin Garrido, Jean Ponce, Xinlei Chen, Michael Rabbat, Yann LeCun, Mido Assran, and Nicolas Ballas.
\newblock V-jepa: Latent video prediction for visual representation learning.
\newblock 2023.

\bibitem{DBLP:journals/corr/abs-2005-14165}
Tom~B. Brown, Benjamin Mann, Nick Ryder, Melanie Subbiah, Jared Kaplan, Prafulla Dhariwal, Arvind Neelakantan, Pranav Shyam, Girish Sastry, Amanda Askell, Sandhini Agarwal, Ariel Herbert{-}Voss, Gretchen Krueger, Tom Henighan, Rewon Child, Aditya Ramesh, Daniel~M. Ziegler, Jeffrey Wu, Clemens Winter, Christopher Hesse, Mark Chen, Eric Sigler, Mateusz Litwin, Scott Gray, Benjamin Chess, Jack Clark, Christopher Berner, Sam McCandlish, Alec Radford, Ilya Sutskever, and Dario Amodei.
\newblock Language models are few-shot learners.
\newblock {\em CoRR}, abs/2005.14165, 2020.

\bibitem{DINO}
Mathilde Caron, Hugo Touvron, Ishan Misra, Herv{\'{e}} J{\'{e}}gou, Julien Mairal, Piotr Bojanowski, and Armand Joulin.
\newblock Emerging properties in self-supervised vision transformers.
\newblock {\em CoRR}, abs/2104.14294, 2021.

\bibitem{Shapenet}
Angel~X. Chang, Thomas Funkhouser, Leonidas Guibas, Pat Hanrahan, Qixing Huang, Zimo Li, Silvio Savarese, Manolis Savva, Shuran Song, Hao Su, Jianxiong Xiao, Li Yi, and Fisher Yu.
\newblock {ShapeNet: An Information-Rich 3D Model Repository}.
\newblock Technical Report arXiv:1512.03012 [cs.GR], Stanford University --- Princeton University --- Toyota Technological Institute at Chicago, 2015.

\bibitem{PointGPT}
Guangyan Chen, Meiling Wang, Yi Yang, Kai Yu, Li Yuan, and Yufeng Yue.
\newblock Pointgpt: Auto-regressively generative pre-training from point clouds.
\newblock {\em Advances in Neural Information Processing Systems}, 36, 2024.

\bibitem{SimCLR}
Ting Chen, Simon Kornblith, Mohammad Norouzi, and Geoffrey Hinton.
\newblock A simple framework for contrastive learning of visual representations.
\newblock In Hal~Daumé III and Aarti Singh, editors, {\em Proceedings of the 37th International Conference on Machine Learning}, volume 119 of {\em Proceedings of Machine Learning Research}, pages 1597--1607. PMLR, 13--18 Jul 2020.

\bibitem{devlin2018bert}
Jacob Devlin, Ming{-}Wei Chang, Kenton Lee, and Kristina Toutanova.
\newblock {BERT:} pre-training of deep bidirectional transformers for language understanding.
\newblock {\em CoRR}, abs/1810.04805, 2018.

\bibitem{du2021self}
Bi'an Du, Xiang Gao, Wei Hu, and Xin Li.
\newblock Self-contrastive learning with hard negative sampling for self-supervised point cloud learning.
\newblock In {\em Proceedings of the 29th ACM International Conference on Multimedia}, pages 3133--3142, 2021.

\bibitem{fps}
Y. Eldar, M. Lindenbaum, M. Porat, and Y.Y. Zeevi.
\newblock The farthest point strategy for progressive image sampling.
\newblock In {\em Proceedings of the 12th IAPR International Conference on Pattern Recognition, Vol. 2 - Conference B: Computer Vision \& Image Processing. (Cat. No.94CH3440-5)}, pages 93--97 vol.3, 1994.

\bibitem{fu2022distillation}
Kexue Fu, Peng Gao, Renrui Zhang, Hongsheng Li, Yu Qiao, and Manning Wang.
\newblock Distillation with contrast is all you need for self-supervised point cloud representation learning, 2022.

\bibitem{BYOL}
Jean{-}Bastien Grill, Florian Strub, Florent Altch{\'{e}}, Corentin Tallec, Pierre~H. Richemond, Elena Buchatskaya, Carl Doersch, Bernardo~{\'{A}}vila Pires, Zhaohan~Daniel Guo, Mohammad~Gheshlaghi Azar, Bilal Piot, Koray Kavukcuoglu, R{\'{e}}mi Munos, and Michal Valko.
\newblock Bootstrap your own latent: {A} new approach to self-supervised learning.
\newblock {\em CoRR}, abs/2006.07733, 2020.

\bibitem{guo2020deep}
Yulan Guo, Hanyun Wang, Qingyong Hu, Hao Liu, Li Liu, and Mohammed Bennamoun.
\newblock Deep learning for 3d point clouds: A survey.
\newblock {\em IEEE transactions on pattern analysis and machine intelligence}, 43(12):4338--4364, 2020.

\bibitem{MAE}
Kaiming He, Xinlei Chen, Saining Xie, Yanghao Li, Piotr Doll{\'{a}}r, and Ross~B. Girshick.
\newblock Masked autoencoders are scalable vision learners.
\newblock {\em CoRR}, abs/2111.06377, 2021.

\bibitem{STRL}
Siyuan Huang, Yichen Xie, Song-Chun Zhu, and Yixin Zhu.
\newblock Spatio-temporal self-supervised representation learning for 3d point clouds, 2021.

\bibitem{jaiswal2020survey}
Ashish Jaiswal, Ashwin~Ramesh Babu, Mohammad~Zaki Zadeh, Debapriya Banerjee, and Fillia Makedon.
\newblock A survey on contrastive self-supervised learning.
\newblock {\em Technologies}, 9(1):2, 2020.

\bibitem{JEPA}
Yann LeCun.
\newblock A path towards autonomous machine intelligence, 2022.

\bibitem{SGDR}
Ilya Loshchilov and Frank Hutter.
\newblock {SGDR:} stochastic gradient descent with restarts.
\newblock {\em CoRR}, abs/1608.03983, 2016.

\bibitem{AdamW}
Ilya Loshchilov and Frank Hutter.
\newblock Fixing weight decay regularization in adam.
\newblock {\em CoRR}, abs/1711.05101, 2017.

\bibitem{mei2024unsupervised}
Guofeng Mei, Cristiano Saltori, Elisa Ricci, Nicu Sebe, Qiang Wu, Jian Zhang, and Fabio Poiesi.
\newblock Unsupervised point cloud representation learning by clustering and neural rendering.
\newblock {\em International Journal of Computer Vision}, pages 1--19, 2024.

\bibitem{morton}
M Morton, G.
\newblock {\em A Computer Oriented Geodetic Data Base and a New Technique in File Sequencing.}
\newblock International Business Machines, 1966.

\bibitem{mu2022slip}
Norman Mu, Alexander Kirillov, David Wagner, and Saining Xie.
\newblock Slip: Self-supervision meets language-image pre-training.
\newblock In {\em European conference on computer vision}, pages 529--544. Springer, 2022.

\bibitem{oord2018representation}
Aaron van~den Oord, Yazhe Li, and Oriol Vinyals.
\newblock Representation learning with contrastive predictive coding.
\newblock {\em arXiv preprint arXiv:1807.03748}, 2018.

\bibitem{Point-MAE}
Yatian Pang, Wenxiao Wang, Francis~EH Tay, Wei Liu, Yonghong Tian, and Li Yuan.
\newblock Masked autoencoders for point cloud self-supervised learning.
\newblock In {\em European conference on computer vision}, pages 604--621. Springer, 2022.

\bibitem{PointNet}
Charles~R. Qi, Hao Su, Kaichun Mo, and Leonidas~J. Guibas.
\newblock Pointnet: Deep learning on point sets for 3d classification and segmentation, 2017.

\bibitem{PointNet++}
Charles~Ruizhongtai Qi, Li Yi, Hao Su, and Leonidas~J. Guibas.
\newblock Pointnet++: Deep hierarchical feature learning on point sets in a metric space.
\newblock {\em CoRR}, abs/1706.02413, 2017.

\bibitem{radford2019language}
Alec Radford, Jeffrey Wu, Rewon Child, David Luan, Dario Amodei, Ilya Sutskever, et~al.
\newblock Language models are unsupervised multitask learners.
\newblock {\em OpenAI blog}, 1(8):9, 2019.

\bibitem{DBLP:journals/corr/abs-1910-10683}
Colin Raffel, Noam Shazeer, Adam Roberts, Katherine Lee, Sharan Narang, Michael Matena, Yanqi Zhou, Wei Li, and Peter~J. Liu.
\newblock Exploring the limits of transfer learning with a unified text-to-text transformer.
\newblock {\em CoRR}, abs/1910.10683, 2019.

\bibitem{Saunder2021}
Jonathan Sauder and Bjarne Sievers.
\newblock Context prediction for unsupervised deep learning on point clouds.
\newblock {\em CoRR}, abs/1901.08396, 2019.

\bibitem{hilbert-program}
John Skilling.
\newblock Programming the hilbert curve.
\newblock volume 707, pages 381--387, 04 2004.

\bibitem{ScanObj}
Mikaela~Angelina Uy, Quang{-}Hieu Pham, Binh{-}Son Hua, Duc~Thanh Nguyen, and Sai{-}Kit Yeung.
\newblock Revisiting point cloud classification: {A} new benchmark dataset and classification model on real-world data.
\newblock {\em CoRR}, abs/1908.04616, 2019.

\bibitem{tsne}
Laurens van~der Maaten and Geoffrey~E. Hinton.
\newblock Visualizing data using t-sne.
\newblock {\em Journal of Machine Learning Research}, 9:2579--2605, 2008.

\bibitem{Transformer}
Ashish Vaswani, Noam Shazeer, Niki Parmar, Jakob Uszkoreit, Llion Jones, Aidan~N. Gomez, Lukasz Kaiser, and Illia Polosukhin.
\newblock Attention is all you need.
\newblock {\em CoRR}, abs/1706.03762, 2017.

\bibitem{wang2022self}
Yi Wang, Conrad~M Albrecht, Nassim Ait~Ali Braham, Lichao Mou, and Xiao~Xiang Zhu.
\newblock Self-supervised learning in remote sensing: A review.
\newblock {\em IEEE Geoscience and Remote Sensing Magazine}, 10(4):213--247, 2022.

\bibitem{ModelNet}
Zhirong Wu, Shuran Song, Aditya Khosla, Fisher Yu, Linguang Zhang, Xiaoou Tang, and Jianxiong Xiao.
\newblock 3d shapenets: A deep representation for volumetric shapes.
\newblock In {\em 2015 IEEE Conference on Computer Vision and Pattern Recognition (CVPR)}, pages 1912--1920, 2015.

\bibitem{Point-bert}
Xumin Yu, Lulu Tang, Yongming Rao, Tiejun Huang, Jie Zhou, and Jiwen Lu.
\newblock Point-bert: Pre-training 3d point cloud transformers with masked point modeling.
\newblock In {\em Proceedings of the IEEE/CVF conference on computer vision and pattern recognition}, pages 19313--19322, 2022.

\bibitem{Point2Vec}
Karim~Abou Zeid, Jonas Schult, Alexander Hermans, and Bastian Leibe.
\newblock Point2vec for self-supervised representation learning on point clouds.
\newblock In {\em DAGM German Conference on Pattern Recognition}, pages 131--146. Springer, 2023.

\bibitem{Point-M2AE}
Renrui Zhang, Ziyu Guo, Rongyao Fang, Bin Zhao, Dong Wang, Yu Qiao, Hongsheng Li, and Peng Gao.
\newblock Point-m2ae: Multi-scale masked autoencoders for hierarchical point cloud pre-training, 2022.

\bibitem{I2P-MAE}
Renrui Zhang, Liuhui Wang, Yu Qiao, Peng Gao, and Hongsheng Li.
\newblock Learning 3d representations from 2d pre-trained models via image-to-point masked autoencoders, 2022.

\bibitem{3DCapsNet}
Yongheng Zhao, Tolga Birdal, Haowen Deng, and Federico Tombari.
\newblock 3d point-capsule networks.
\newblock {\em CoRR}, abs/1812.10775, 2018.

\bibitem{zheng2024point}
Xiao Zheng, Xiaoshui Huang, Guofeng Mei, Yuenan Hou, Zhaoyang Lyu, Bo Dai, Wanli Ouyang, and Yongshun Gong.
\newblock Point cloud pre-training with diffusion models.
\newblock In {\em Proceedings of the IEEE/CVF Conference on Computer Vision and Pattern Recognition}, pages 22935--22945, 2024.

\end{thebibliography}
}

\clearpage
\setcounter{section}{0} 
\renewcommand*{\thesection}{\Alph{section}}
 \maketitlesupplementary

\section{Further Pre-training Details}
\label{pretrain}
\paragraph{Optimization}
We utilize AdamW \cite{AdamW} optimizer with cosine learning decay \cite{SGDR}.
Starting from learning rate of $10^{-5}$, we increase it to $10^{-3}$ in the first 30 epochs and decay it to $10^{-6}$.
The batch size for pretraining is set to 512, and $\beta$ for Smooth L1 loss is set to $2$, similar to Point2Vec \cite{Point2Vec}.
The target encoder and context encoder initially have identical parameters. 
The context encoder's parameters are updated via backpropagation, while the target encoders' parameters are updated using the exponential moving average of the context encoder parameters, that is 
$\overline{\theta} \xleftarrow{} \tau \overline{\theta} + (1 - \tau)\theta$
where $\tau \in [0,1]$ denotes the decay rate. 
We gradually increase the decay rate of the exponential moving average from 0.995 to 1.0 during pretraining.

\paragraph{Masking and Ordering}
To determine the sequence of patch embeddings, we utilize the iterative ordering of associated center points, as previously mentioned.
We chose the starting point in this sequence with the lowest sum of its coordinates.
This method allows us to start the sequence from a point on the outer edge of the object rather than from a point within the object's interior.
This consistency in selecting the initial point is experimentally shown to deliver a slightly better learned representation than taking the first available index.

For masking, we define a range of ratios with both upper and lower limits similar to I-JEPA \cite{I-JEPA}.
To start with, we clarify that the term \enquote{block} refers to a sequence of patch embeddings and their corresponding encoded embeddings that are contiguous. 
Because of the sequencing process applied before the target and context selection, most contiguous patch embeddings and encoded embeddings are also spatially contiguous.
For the target, we randomly select 4 blocks of encoded embeddings processed by transformer blocks from within the 0.15 to 0.2 range.
We then remove the corresponding patch embeddings of encoded embedding vectors that have already been chosen as targets for further selection.
Following this, we choose a block of patch embeddings that is within the range of 0.4 to 0.75 out of available patch embeddings that are not concealed. 
Because some of the patch embeddings are not available for context selection, we note that context block usually consists of multiple sets of patch embeddings that are spatially contiguous.
The selection of targets is completed on a per-batch basis, and we track the indices of these targets to ensure that the corresponding patch embeddings of these selected encoded embeddings are concealed in the context selection.
The context is then selected using the available indices of patch embeddings also on a per-batch basis.
\begin{table}[h]
        \centering
        \normalsize{
            \begin{tabular}{l l l c}
            \toprule
            \multicolumn{2}{l}{\bf Targets} & \bf Context & \bf OA \\
            \cmidrule(lr){1-2}\cmidrule(lr){3-3}\cmidrule(lr){4-4}
            Ratio & Freq. & Ratio &  Modelnet40 Linear \\
            \toprule
            (0.1, 0.2) &  4 & (0.85, 1.0) & 93.0 \\
            (0.15, 0.2) &  4 & (0.85, 1.0) &  \textbf{93.3} \\
            (0.2, 0.25) &  4 & (0.85, 1.0) & 93.2 \\
            (0.25, 0.3) &  4 & (0.85, 1.0) &  92.4 \\
            (0.3, 0.35) &  4 & (0.85, 1.0) &  90.5 \\
            (0.35, 0.4) &  4 & (0.85, 1.0) & 84.6 \\
                        \bottomrule

            \end{tabular}
        }
        \caption[Ratio of Targets]{
            \textbf{Ratio Range for Target.}
            The ratio of encoded embedding vectors selected for each target.
        }
        \label{tab:target_ratio}
\end{table}

\section{Further Ablation}
\paragraph{Ratio of Targets.}
\label{seq:ration_tg}
We change the ratio of the selected embedding vectors for the target selection while keeping the number of target blocks and the ratio of context patch embedding fixed.
As shown in \cref{tab:target_ratio}, the performance increases when you increase the ratio to a certain point. 
However, beyond this point, further increasing the ratio results in decreased performance.
This implies that Point-JEPA does not require a large size for the target blocks and benefits from a sufficient amount of available patch embeddings for context selection.

\begin{table}[h]
        \centering
        \normalsize{
            \begin{tabular}{l l l c}
            \toprule
            \multicolumn{2}{l}{\bf Targets} & \bf Context & \bf OA \\
            \cmidrule(lr){1-2}\cmidrule(lr){3-3}\cmidrule(lr){4-4}
            Ratio & Freq. & Ratio & Modelnet40 Linear \\
            \toprule
            (0.15, 0.2) &  4 & (0.85, 1.0) & 93.1 \\
            (0.15, 0.2) &  4 & (0.75, 1.0) &  92.8 \\
            (0.15, 0.2) &  4 & (0.65, 1.0) & 93.4 \\
            (0.15, 0.2) &  4 & (0.45, 1.0) &  93.6 \\
            (0.15, 0.2) &  4 & (0.6, 0.75) &  93.4 \\
            (0.15, 0.2) &  4 & (0.5, 0.75) & 93.1 \\
            (0.15, 0.2) &  4 & (0.4, 0.75) & \textbf{93.7} \\
                        \bottomrule

            \end{tabular}
        }
        \caption[Ratio of Context]{
        \textbf{Ratio Range for Context.}
        The ratio of patch embeddings selected for context encoding.}
        \label{tab:context_ratio}
\end{table}
\begin{figure*}[t]
    \centering
    \begin{subfigure}[b]{0.45\textwidth}
        \centering
        \includegraphics[width=\textwidth]{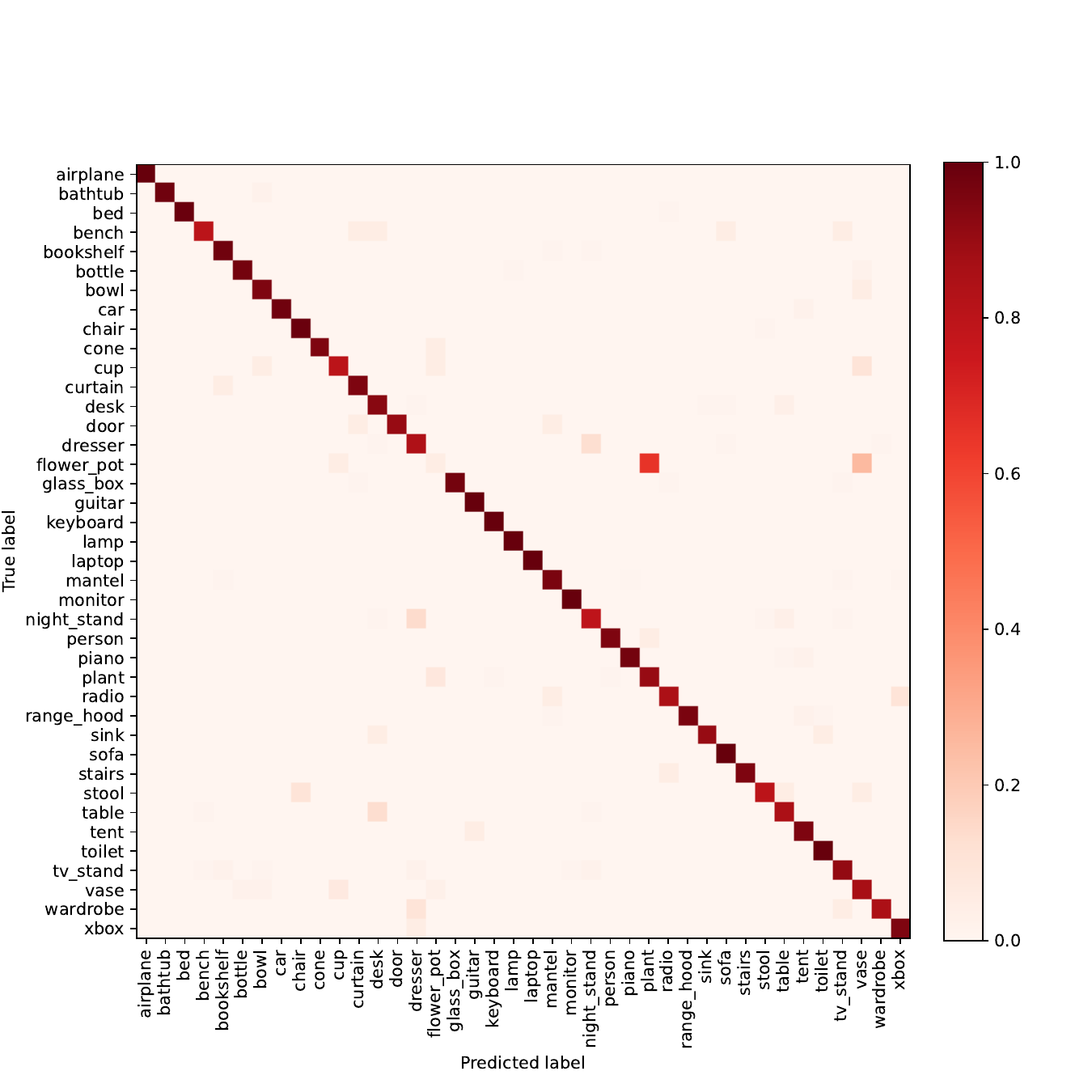}
        \caption{Row-normalized confusion matrix on ModelNet40}
        \label{fig:row_norm}
    \end{subfigure}
    \begin{subfigure}[b]{0.45\textwidth}
        \centering
        \includegraphics[width=\textwidth]{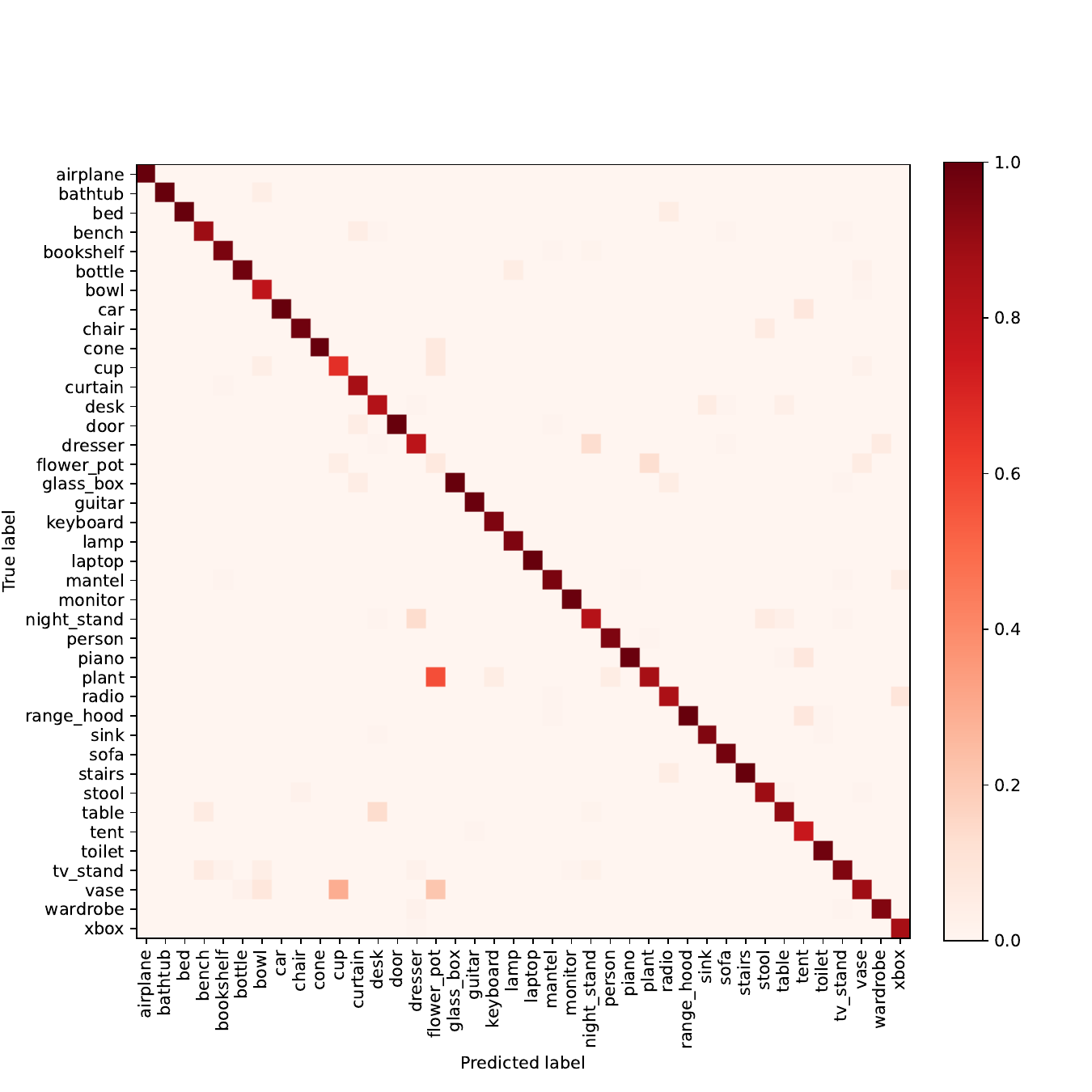}
        \caption{Column-normalized confusion matrix on ModelNet40}
        \label{fig:col_norm}
    \end{subfigure}
     \begin{subfigure}[b]{0.45\textwidth}
        \centering
        \includegraphics[width=\textwidth]{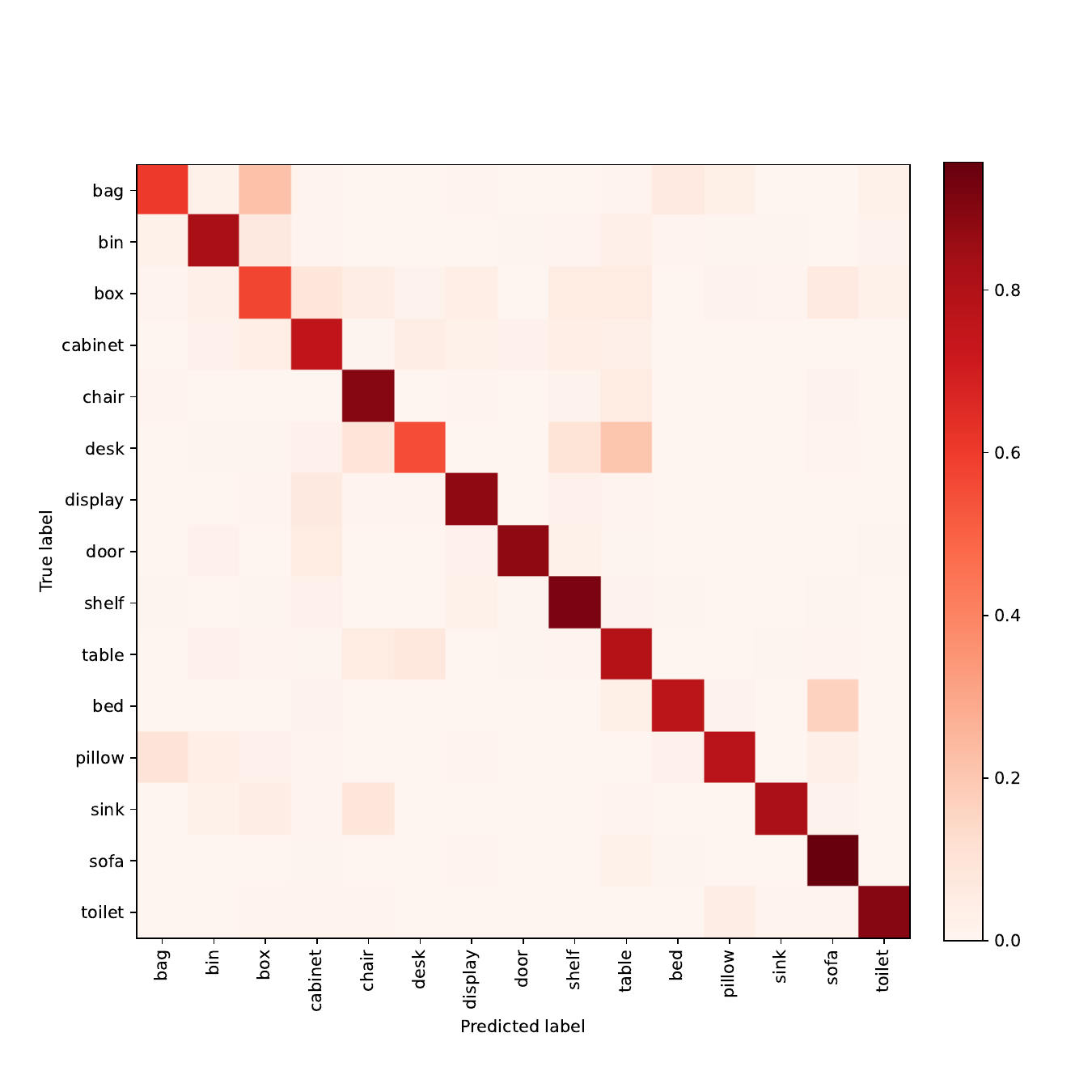}
        \caption{Row-normalized confusion matrix on ScanObjNN}
        \label{fig:col_norm}
    \end{subfigure}
      \begin{subfigure}[b]{0.45\textwidth}
        \centering
        \includegraphics[width=\textwidth]{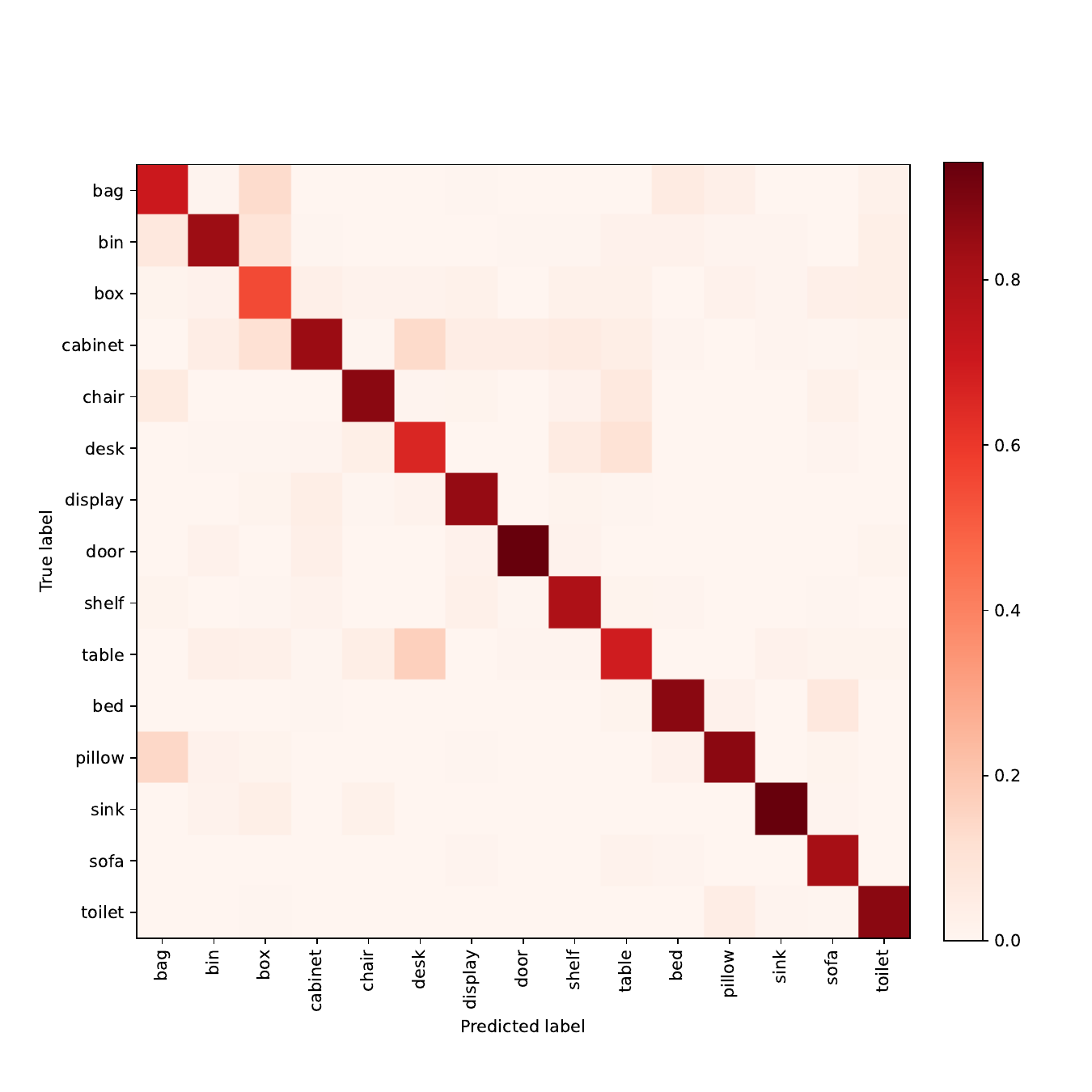}
        \caption{Column-normalized confusion matrix on ScanObjNN}
        \label{fig:col_norm}
    \end{subfigure}

    \caption{Confusion matrices illustrating model performance on ModelNet40 and another dataset, highlighting class-specific accuracies and challenges with similar categories.}
    \label{fig:confusion_matrix}
\end{figure*}
\paragraph{Ratio of Context.}
In this study, we change the ratio of patch embeddings selected for context encoding while keeping the number of targets and the ratio range for targets fixed.
As shown in Table  \ref{tab:context_ratio}, having a relatively large difference between the lower and upper bound of the ratio can improve performance.
In other words, Point-JEPA learns a better representation when the number of selected context patch embeddings varies more between training iterations.
Additionally, when the upper bound of the ratio is somewhat constrained, we see increased performance.

\paragraph{Predictor Depth}
We also study the effect of the predictor's depth on the learned representation.
To this end, we vary the predictor depth and observe its effect on the linear evaluation accuracy.
As shown in Table \ref{tab:predictor_depth}, Point-JEPA benefits from a deeper predictor.

\begin{table}[htbp]
        \centering
        \normalsize{
            \begin{tabular}{l c}
            \toprule
            Predictor Depth &  Modelnet40 Linear (OA) \\
            \midrule
            2 & 92.5\\
            3 & 92.8 \\
            4 & 93.2\\
            5 & 93.4\\
            6 & \textbf{93.7} \\
            \bottomrule
            \end{tabular}
        }
        \caption[Ratio of patch embeddings for target encoding]{
        \textbf{Predictor Depth.}
        Predictor depth and its effect on learned representation.
        }
        \label{tab:predictor_depth}
\end{table}

\paragraph{Class confusion on ModelNet40 and ScanObjNN}
To assess our model's performance on the ModelNet40 \cite{ModelNet} and ScanObjNN \cite{ScanObj} datasets, we present two types of visualizations for each dataset. The first is a row-normalized confusion matrix, which illustrates the model's sensitivity, indicating how well the model identifies each actual class. The second is a column-normalized confusion matrix, depicting the model's specificity, which shows the correctness of predictions for each class assumed by the model.
As illustrated in parts (a) and (b) of \cref{fig:confusion_matrix}, the model fine-tuned on ModelNet40 demonstrates high accuracy. At the same time, errors predominantly arise from similar categories within the dataset. For instance, \enquote{flower pot} and \enquote{plant} are often misclassified, likely due to the presence of flowers in some of the flower pot models in the ModelNet40 dataset.
Similarly, parts (c) and (d) of \cref{fig:confusion_matrix} show the aforementioned confusion matrices. As highlighted in the main paper, our
model’s performance on ScanObjectNN dataset has room
for enhancement compared to ModelNet40. The confusion
matrix reveals some misclassifications, but it is encouraging
to see that these errors predominantly occur between
closely related classes, such as ‘table’ and ‘desk’ or ‘sofa’
and ‘bed’. This suggests that our model has a solid grasp of
the key characteristics of these categories and that further
refinement of the classification criteria could lead to significant
improvements in overall accuracy.

\end{document}